\documentclass[times,review,10pt]{elsarticle}

\usepackage{amssymb}
\usepackage{amsmath}
\usepackage{url}
\usepackage{mathtools}
\usepackage{booktabs}
\usepackage{makecell}

\journal{Pattern Recognition}

\begin{document}

\begin{frontmatter}



\title{AuricularWorld: Hierarchical Action-Guided World Modeling for Fine-Grained Auricular Structure Segmentation from CT Scans}


\author[inst1]{Jingwen Yang\cormark[1]} 
\author[inst1]{Senmao Wang\cormark[1]} 
\author[inst2]{Luoyao Kang\cormark[1]}
\author[inst1]{Runmeng Cui}
\author[inst1]{Keying Zhang}
\author[inst1]{Yunjia Bao}
\author[inst3]{Haifan Gong\cormark[2]}
\ead{haifangong@link.cuhk.edu.cn}
\author[inst1]{Lin Lin\cormark[2]}
\ead{linlin@psh.pumc.edu.cn}
\author[inst1]{Haiyue Jiang\cormark[2]}
\ead{haiyuejiang65@sina.com}

\cortext[1]{These authors contributed equally to this work.}

\cortext[2]{Corresponding authors.}

\affiliation[inst1]{
            organization={Plastic Surgery Hospital, Chinese Academy of Medical Sciences \& Peking Union Medical College},
            city={Beijing},
            postcode={100144}, 
            country={China}}

\affiliation[inst2]{
            organization={The Chinese University of Hong Kong},
            city={Hong Kong SAR},
            postcode={999077}, 
            country={China}}

\affiliation[inst3]{
            organization={The Chinese University of Hong Kong (Shenzhen)},
            city={Shenzhen},
            postcode={518172}, 
            country={China}}

\begin{abstract}
Fine-grained segmentation of auricular structures in CT is challenging due to the small size of the ear region, highly irregular cartilage boundaries, and ambiguous interfaces between cartilage and surrounding soft tissues. Moreover, clinical annotations often contain both composite anatomical structures with cartilage and adjacent skin and their corresponding cartilage-only regions, resulting in inherently nested and overlapping labels.
In this work, we propose a world-model-based segmentation framework that enables iterative anatomical reasoning beyond conventional feed-forward prediction. Built upon an encoder–decoder segmentation architecture, our framework integrates a deterministic recurrent state-space model into the intermediate latent space. Multi-scale encoder features and partially decoded representations are fused to construct a structural observation that initializes the latent dynamics. During inference, the model performs a three-step latent rollout without ground-truth guidance, where hierarchical anatomical actions are predicted to update the recurrent state and evolve the latent representation. The refined latent trajectory is subsequently projected back to the decoder and combined with high-resolution features to generate the final segmentation. To facilitate learning of reliable latent transitions, we introduce a balanced hierarchical action objective that addresses foreground sparsity, missing anatomical groups, and the imbalance between add and remove operations.
Extensive experiments demonstrate that the proposed framework consistently improves segmentation accuracy and reduces the HD95 boundary error for over 43\% for small, irregular, and overlapping auricular structures in CT, validating the effectiveness of latent world-model reasoning for challenging medical image segmentation.
\end{abstract}




\end{frontmatter}

\section{Introduction}
\label{sec:introduction}

Accurate segmentation of fine-grained auricular anatomy from computed tomography (CT) is important for quantitative anatomical analysis, patient-specific modeling, and reconstructive surgical planning~\cite{rodriguezarias2022microtia,mohamed2023segmentation}. Yet this task remains particularly challenging. The external ear occupies only a small portion of a whole-head CT volume, leading to severe foreground--background imbalance, as commonly encountered in fine-grained head-and-neck segmentation~\cite{zhu2019anatomynet}. Moreover, auricular cartilage exhibits thin, irregular geometries and weak contrast against surrounding soft tissues, making its boundaries difficult to delineate reliably~\cite{wang2026costal}. The problem is further complicated by the coexistence of skin-covered auricular subunits and their corresponding cartilage-only regions, which naturally form nested and partially overlapping anatomical labels. Consequently, successful segmentation requires not only local boundary precision, but also a coherent representation of the structural relationships among closely coupled anatomical regions.

Deep learning has substantially advanced medical image analysis~\cite{kang2023visual}, especially in the field of segmentation~\cite{isensee2021nnunet,gong2021multitask,gong2023thyroid,gong2024intensity,xu2023asc,huang2025bcnet,gong2025boundary,wang2026costal}. U-Net~\cite{ronneberger2015unet} established the encoder--decoder paradigm with skip connections, while nnU-Net~\cite{isensee2021nnunet} demonstrated that carefully configured convolutional pipelines can generalize robustly across diverse segmentation tasks. Transformer-based architectures further improve long-range contextual modeling~\cite{chen2021transunet,zhou2023nnformer}, and recent state-space models provide an efficient alternative for modeling volumetric dependencies~\cite{gong2025nnmamba}. Despite these advances, most existing approaches retain the same fundamental formulation: an image is mapped through a feed-forward network to a final segmentation mask. Intermediate representations are optimized primarily for direct voxel-wise prediction, rather than explicitly modeling how an incomplete anatomical hypothesis should evolve toward a structurally consistent solution.

We argue that fine-grained auricular segmentation can instead be formulated as a \emph{latent anatomical evolution process}. Rather than requiring a network to infer the complete anatomy in a single feed-forward transformation, an intermediate anatomical state can be progressively updated through semantic corrections. This perspective is closely related to world models, which learn compact latent states together with transition dynamics describing how these states evolve under actions~\cite{ha2018worldmodels}. In our setting, however, the latent state represents an anatomical hypothesis rather than an external environment, while an action represents an internal semantic correction that modifies the current representation toward a more anatomically plausible state.

Based on this formulation, we propose \emph{AuricularWorld}, an action-conditioned latent world model for fine-grained auricular CT segmentation. AuricularWorld embeds recurrent latent dynamics into an encoder--decoder segmentation architecture. Multi-scale encoder features and partially decoded representations are integrated to construct an anatomical observation, from which the latent state is initialized. The model then performs a short sequence of action-conditioned latent transitions to progressively refine this representation before returning it to the high-resolution decoder. Importantly, the world-model rollout is performed entirely in feature space rather than directly on segmentation masks. This design allows latent dynamics to capture structural evolution while retaining skip connections and high-resolution decoding for precise boundary reconstruction.

A key component of AuricularWorld is the explicit definition of \emph{hierarchical anatomical actions}. These actions describe semantic addition and removal operations between successive anatomical states, providing structured supervision for latent transitions. Learning such actions is non-trivial because patch-based auricular segmentation contains substantial background, frequently absent anatomical groups, and highly asymmetric transition frequencies. We therefore introduce a foreground-masked balanced action objective that suppresses uninformative background supervision and compensates for imbalanced anatomical transitions, enabling the latent dynamics to focus on meaningful structural changes.

To support systematic evaluation, we further construct a fine-grained auricular CT dataset comprising 193 patient examinations and 198 annotated auricles. The dataset contains complementary annotations of skin-covered auricular subunits and their corresponding cartilage structures, established through clinical annotation and expert consensus review. Extensive experiments demonstrate that the proposed latent-evolution formulation consistently improves the segmentation of small, irregular, and structurally overlapping auricular regions.

The main contributions of this work are summarized as follows:
\begin{itemize}
\item We establish a fine-grained auricular CT segmentation dataset containing complementary annotations of skin-covered auricular subunits and their corresponding cartilage structures, providing a dedicated benchmark for this clinically relevant yet underexplored task.

\item We formulate fine-grained segmentation as an \emph{action-conditioned latent anatomical evolution process}, moving beyond conventional single-pass image-to-mask prediction by explicitly modeling progressive refinement of anatomical representations.

\item We develop a multi-scale recurrent world model that performs latent rollouts within the feature space while preserving high-resolution decoding pathways, enabling structural refinement without sacrificing boundary localization.

\item We introduce hierarchical anatomical actions together with a foreground-masked balanced objective to learn reliable latent transitions under sparse foreground supervision, missing anatomical groups, and highly asymmetric transition distributions.

\end{itemize}

\section{Related Work}
\subsection{Deep Learning-based Medical Image Segmentation}

Modern medical image segmentation is predominantly built upon encoder--decoder architectures. U-Net~\cite{ronneberger2015unet} introduced skip-connected decoding to combine high-level semantic representations with fine spatial details, while nnU-Net~\cite{isensee2021nnunet} further demonstrated that task-adaptive preprocessing, architecture configuration, training, and post-processing can yield robust performance across heterogeneous biomedical datasets. Recent studies have continued to improve this paradigm through hierarchical feature aggregation and progressive refinement. For example, hierarchical multi-scale feature modeling has been explored to better integrate representations across resolutions~\cite{song2025hierarchical}, while DPGNet introduces progressive boundary-aware refinement under uncertainty~\cite{wang2025dpgnet}. Related hierarchical information integration strategies further emphasize the importance of preserving fine structural details during decoding~\cite{zhang2026carving,yu2026rethinking}. These developments have established encoder--decoder networks as a strong foundation for fine-grained medical image segmentation, as also reflected in recent surveys of modern segmentation architectures~\cite{zhu2026survey}.

To enhance global context modeling beyond local convolutional operations, recent medical image segmentation methods have increasingly adopted Transformer-~\cite{vaswani2017attention} and state-space-based architectures~\cite{gu2024mamba,dao2024transformers}. Transformer approaches such as TransUNet~\cite{chen2021transunet} and nnFormer~\cite{zhou2023nnformer} leverage self-attention to capture long-range spatial dependencies, while more recent Mamba-based methods provide an efficient alternative through selective state-space modeling. For example, nnMamba~\cite{gong2025nnmamba} extends state-space modeling to volumetric medical images, SliceMamba~\cite{fan2025slicemamba} combines Mamba representations with neural architecture search, and GraphMamba~\cite{yu2026graphmamba} further incorporates graph-driven spatial ordering to model anatomical relationships. Recent analyses also demonstrate the effectiveness of Mamba architectures for 3D volumetric segmentation across diverse medical imaging datasets~\cite{wang2026comprehensive}. Despite their improved capability to capture global context and long-range dependencies, both Transformer- and Mamba-based approaches primarily treat attention or state transitions as feature-propagation mechanisms within a feed-forward segmentation pipeline, rather than explicitly modeling semantically meaningful anatomical transitions that progressively evolve the latent representation.

Another line of research focuses on incorporating anatomical organization and boundary information into segmentation. Hierarchy-aware segmentation explicitly models dependencies among anatomically related or nested labels~\cite{cheng2026deep}, which is particularly relevant when different annotations correspond to overlapping structures at different levels of granularity. Boundary-focused approaches improve discrimination around uncertain interfaces through contrastive or alignment-based supervision~\cite{yang2025boundary,huang2026boundary}, while multimodal boundary-aware fusion has also been explored for structures with heterogeneous appearance~\cite{zhou2025boundary}. More generally, anatomical priors, boundary representations, appearance consistency, and structure-aware supervision have been shown to improve medical image segmentation and landmark localization~\cite{gong2021multitask,gong2023thyroid,gong2024intensity,xu2023asc,gong2025boundary,gong2025fetal,wang2026costal}. These strategies are particularly valuable for fine-grained structures whose boundaries are weak, irregular, or spatially constrained by neighboring anatomy.

Despite substantial progress in feature extraction, global-context modeling, and structure-aware supervision, existing methods predominantly optimize a direct mapping from an image to a segmentation mask. Even when recurrent, Transformer, or state-space operations are employed, intermediate representations typically evolve implicitly through stacked computational layers, without explicitly defining \emph{what anatomical transition should occur} at each refinement step. This distinction is particularly important for fine-grained auricular CT segmentation, where the ear occupies only a small portion of the volume, cartilage boundaries are highly irregular and poorly contrasted, and composite auricular structures can spatially overlap with their corresponding cartilage-only regions. AuricularWorld addresses this gap by explicitly formulating segmentation as an iterative latent reasoning process: hierarchical anatomical actions condition state transitions during a multi-step rollout, progressively evolving the structural representation before it is projected back to the decoder for final segmentation.

\subsection{Auricular Structure Analysis and Segmentation}

Automated auricular analysis has attracted increasing interest for morphological assessment, personalized modeling, and reconstructive surgery. Existing studies have predominantly focused on surface-visible anatomy. Mussi et al.~\cite{mussi2021ear} proposed an image-processing approach for identifying auricular elements from depth maps, while Servi et al.~\cite{servi2021auricular} explored U-Net-based segmentation of auricular substructures using similar surface representations. These approaches demonstrate the feasibility of automated auricular decomposition, but they primarily characterize external morphology and therefore cannot directly resolve internal cartilage anatomy from volumetric clinical images.

Patient-specific auricular reconstruction has also benefited from three-dimensional acquisition and modeling. Wang et al.~\cite{wang2022auricular} investigated 3D auricular subunit models for cartilage framework fabrication, Rodríguez-Arias et al.~\cite{rodriguezarias2022microtia} developed a segmentation protocol for patient-specific 3D models in microtia reconstruction, and Ross et al.~\cite{ross2022ultrasound} investigated ultrasound-derived 3D models for personalized auricular implants. Although these studies highlight the clinical value of patient-specific auricular modeling, anatomical extraction is generally embedded within a reconstruction workflow and often involves manual or semi-automatic processing.

In contrast, our task requires fully automatic delineation of multiple fine-grained auricular subunits and their corresponding cartilage regions directly from CT volumes. The combination of extremely small target regions, weak cartilage--soft-tissue contrast, irregular boundaries, and nested skin--cartilage relationships makes this problem fundamentally different from conventional surface-based ear analysis. These characteristics motivate both a dedicated clinically annotated benchmark and a segmentation framework capable of reasoning about structural evolution rather than relying solely on direct voxel-wise prediction.

\subsection{World Models and Latent Dynamics Learning}

World models aim to learn compact latent representations together with transition dynamics that predict how internal states evolve under actions~\cite{ha2018worldmodels}. PlaNet~\cite{hafner2019planet} introduced the recurrent state-space model (RSSM), combining recurrent deterministic states with stochastic latent variables to support multi-step prediction in latent space. Dreamer~\cite{hafner2020dreamer} further demonstrated that behaviors can be learned from imagined latent trajectories without repeatedly interacting with the external environment. Together, these studies established latent rollout as an effective mechanism for representing state evolution and evaluating action-conditioned transitions.

The world-model paradigm has recently begun to extend beyond conventional control settings. In medical imaging~\cite{chen2026medicalworldmodels}, DreamReg~\cite{kang2026dreamreg} employs belief-driven world modeling for 2D--3D ultrasound registration. CheXWorld~\cite{yue2025chexworld} models local anatomy, global anatomical organization, and imaging-domain variation to improve self-supervised radiographic representation learning. Medical World Model~\cite{yang2025medical} instead models treatment-conditioned tumor evolution for clinical simulation and treatment planning. In parallel, biomedical agents are increasingly being developed for sequential clinical reasoning, multimodal interaction, and specialized tool use~\cite{moritz2025coordinated,kong2026dEnantiomeric,schmidgall2026agentclinic}. These studies demonstrate that latent dynamics can provide useful inductive biases for medical image representation and longitudinal prediction.

Our work differs from these approaches in both the meaning of the latent state and the role of the action. We do not model physical interaction, longitudinal disease progression, future-image generation, or reward-based planning. Instead, the latent state represents an intermediate anatomical hypothesis, and each action specifies a semantic correction to that hypothesis. An RSSM-based module performs a short action-conditioned rollout entirely within the segmentation feature space, after which the refined representation is returned to the high-resolution decoder. Thus, AuricularWorld introduces latent dynamics as an explicit mechanism for \emph{anatomical representation refinement}, providing a complementary formulation to both conventional feed-forward segmentation networks and existing medical world models.

\begin{table}[t]
    \centering
    \caption{Summary and patient-level partition of the proposed auricular CT dataset.}
    \label{tab:dataset_summary}
    \small
    \setlength{\tabcolsep}{5pt}
    \renewcommand{\arraystretch}{1.15}
    \begin{tabular}{@{}lcccc@{}}
        \toprule
        & \textbf{Train} & \textbf{Validation} & \textbf{Test} & \textbf{Total} \\
        \midrule
        Patients & 117 & 41 & 32 & 185 \\
        Annotated auricles & 121 & 41 & 32 & 194 \\
        \midrule
        \multicolumn{5}{@{}l}{
        CT scanners: Philips CT 6000 / Brilliance 64} \\
        \multicolumn{5}{@{}l}{
        Annotation: skin-covered subunits and corresponding cartilage structures} \\
        \multicolumn{5}{@{}l}{
        Labels: 35 atomic foreground labels / canonical structures} \\
        \bottomrule
    \end{tabular}
\end{table}

\section{Dataset}

\subsection{Dataset Motivation}

Large-scale annotated datasets have substantially advanced medical image segmentation. Representative benchmarks such as WORD~\cite{luo2022word} and AMOS~\cite{ji2022amos} focus on abdominal multi-organ segmentation, while TotalSegmentator~\cite{wasserthal2023totalsegmentator} provides annotations for a broad range of anatomical structures in CT. In head-and-neck imaging, datasets such as HaN-Seg~\cite{podobnik2023hanseg} and AnatomyNet~\cite{zhu2019anatomynet} mainly target organs at risk for radiotherapy planning. Despite their importance, these datasets primarily characterize major organs or clinically defined structures and do not provide fine-grained annotations of auricular anatomy. Existing auricular studies are predominantly based on surface or depth-map representations~\cite{servi2021auricular}, which cannot directly characterize the volumetric relationship between external auricular morphology and the underlying cartilage.

This limitation is particularly relevant to auricular reconstruction, where both external morphology and the geometry of the supporting cartilage framework are clinically important. Auricular CT segmentation is further complicated by the small spatial extent of the ear, weak cartilage--soft-tissue contrast, irregular boundaries, and nested relationships between skin-covered subunits and their corresponding cartilage components. We therefore establish a dedicated fine-grained auricular CT dataset containing complementary annotations of external auricular subunits and cartilage structures. Unlike conventional multi-organ benchmarks, the proposed dataset explicitly captures these coupled and partially overlapping anatomical relationships, providing a dedicated benchmark for fine-grained auricular segmentation.

\subsection{Dataset Construction and Annotation}

A total of 193 CT examinations from 193 patients aged (>5) years were retrospectively collected at the Plastic Surgery Hospital, Chinese Academy of Medical Sciences and Peking Union Medical College, Beijing, China, between April 2017 and September 2025. All patients underwent evaluation or treatment for auricular deformities. Most annotated samples corresponded to structurally unaffected contralateral ears from patients with unilateral auricular deformities. One auricle was annotated for 188 patients, while both auricles were included for five patients, resulting in 198 annotated auricles.

Auricles were included when the complete external ear was covered by the CT acquisition and the relevant anatomical boundaries could be reliably identified. Examinations with severe motion artifacts, metal artifacts, or insufficient image quality were excluded. Individual auricles were further excluded when they were incompletely visualized or unsuitable for reliable subunit annotation.

Manual segmentation was performed using 3D Slicer. For each auricle, two complementary annotation sets were generated: skin-covered auricular subunits and their corresponding cartilage structures. Segmentation was performed jointly in the axial, coronal, and sagittal planes and further inspected in three-dimensional view. Three clinicians experienced in auricular reconstruction participated in the annotation and review procedure, and all labels were subsequently checked and corrected by a senior auricular reconstruction specialist to obtain the final consensus annotations.

The original clinical annotations contain nested and partially overlapping anatomical structures. To enable conventional multi-class segmentation, these annotations were decomposed into 35 mutually exclusive foreground \emph{atomic labels} together with background. Parent residual regions were explicitly retained so that the original anatomical hierarchy could be recovered. During evaluation, atomic predictions were deterministically recomposed into 35 canonical anatomical structures using a predefined atomic-to-canonical mapping.

Table~\ref{tab:dataset_summary} summarizes the main characteristics of the proposed dataset. CT examinations were acquired using two Philips scanners. Among the 193 examinations, 162 were acquired using the Philips CT 6000 and 31 using the Philips Brilliance 64. All scans were acquired at 120~kVp with a reconstruction matrix of (512$\times$512). The in-plane pixel spacing ranged from 0.373 to 0.516~mm, while reconstructed slice thickness and interslice spacing ranged from 0.33 to 0.60~mm.

Written informed consent for the use of imaging data was obtained from all participants or, for participants younger than 18 years, from their legal guardians. The study was approved by the Ethics Committee of the Plastic Surgery Hospital, Chinese Academy of Medical Sciences and Peking Union Medical College, Beijing, China. All CT images and associated metadata were anonymized before annotation and analysis.
Dataset partitioning was performed strictly at the patient level. For patients with bilateral annotations, both auricles were assigned to the same subset to prevent patient-level information leakage between training, validation, and test sets. This dataset will be made publicly available after acceptance.

\begin{figure*}[t]
    \centering
    \includegraphics[width=\textwidth]{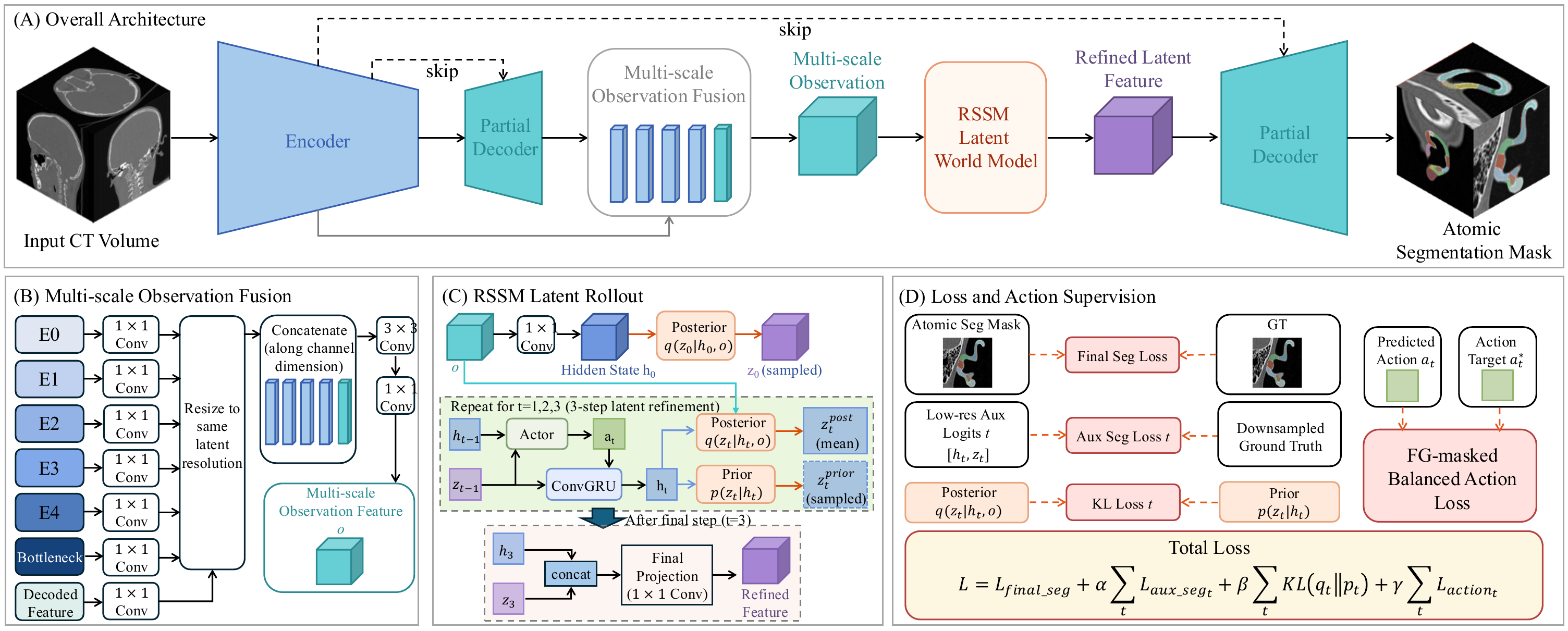}
    \caption{
    Overview of the proposed \textbf{AuricularWorld} framework.
    \textbf{(A) Overall architecture.}
    The input 3D CT volume is processed by the nnU-Net encoder and the first
    part of the decoder to obtain an intermediate decoded feature. Multi-scale
    encoder representations and the decoded feature are fused into an
    observation feature, which is refined by the RSSM latent world model.
    The resulting latent feature is passed to the remaining decoder stages
    together with the high-resolution skip connections to produce the atomic
    segmentation mask.
    \textbf{(B) Multi-scale observation fusion.}
    Features from encoder stages $E_0$--$E_4$, the bottleneck, and the
    partially decoded representation are projected using
    $1\times1\times1$ convolutions, resized to a common latent resolution,
    concatenated along the channel dimension, and fused to obtain the
    multi-scale observation $o$.
    \textbf{(C) RSSM latent rollout.}
    The observation initializes the latent state, after which the actor,
    ConvGRU, posterior, and prior recurrently update the deterministic and
    stochastic states over three refinement steps. Following the final
    transition, the deterministic and stochastic states are concatenated and
    projected to generate the refined latent feature.
    \textbf{(D) Loss and action supervision.}
    AuricularWorld is jointly optimized using the final full-resolution
    segmentation loss, recurrent auxiliary segmentation losses,
    KL-divergence regularization, and foreground-masked balanced action losses.
    }
    \label{fig:framework}
\end{figure*}

\section{Methodology}
\label{sec:methodology}

\subsection{Overview}
\label{subsec:method_overview}

Given an input CT patch
$\mathbf{x}\in\mathbb{R}^{1\times H\times W\times D}$
and its atomic label map
$\mathbf{y}\in\{0,\ldots,C-1\}^{H\times W\times D}$,
where $C$ denotes the number of atomic classes including background, our goal
is to learn a segmentation model $f_{\theta}$ that predicts the corresponding
high-resolution atomic segmentation
$\widehat{\mathbf{y}}=f_{\theta}(\mathbf{x})$.
The nested clinical annotations are converted into mutually exclusive atomic
labels for training and deterministically reconstructed into the canonical
auricular structures after inference. The same label representation is used
for all evaluated methods.

As illustrated in Fig.~\ref{fig:framework}(A), AuricularWorld is built upon
the nnU-Net encoder--decoder architecture~\cite{isensee2021nnunet} and inserts
a recurrent latent world model at an intermediate decoder resolution. Given
$\mathbf{x}$, the encoder produces a hierarchy of multi-scale features
$\mathcal{E}_{\mathbf{x}}$, while the first part of the decoder produces an
intermediate feature $\mathbf{D}_2$ at encoder stage~2. These features are
fused into a multi-scale observation $\mathbf{o}$, which is recurrently
refined by the RSSM world model into a latent representation $\mathbf{r}$.
The remaining high-resolution decoder then combines $\mathbf{r}$ with the
original skip features $\mathbf{E}_1$ and $\mathbf{E}_0$ to generate the final
segmentation:
\begin{equation}
    \mathbf{o}
    = \mathcal{F}_{\mathrm{ms}}
      (\mathcal{E}_{\mathbf{x}},\mathbf{D}_2),
    \qquad
    \mathbf{r}
    = \mathcal{W}(\mathbf{o}),
    \qquad
    \widehat{\mathbf{y}}
    = \mathcal{D}_{\mathrm{high}}
      (\mathbf{r},\mathbf{E}_1,\mathbf{E}_0),
\end{equation}

where $\mathcal{F}_{\mathrm{ms}}$ denotes multi-scale observation fusion,
$\mathcal{W}$ denotes the RSSM latent world model, and
$\mathcal{D}_{\mathrm{high}}$ denotes the remaining high-resolution decoder.
Unlike iterative mask-refinement methods, AuricularWorld recurrently updates
the latent representation rather than the segmentation mask itself, while
preserving the high-resolution decoder and skip connections for detailed
boundary reconstruction.

\subsection{Multi-scale Anatomical Observation}
\label{subsec:observation_fusion}

As shown in Fig.~\ref{fig:framework}(B), the encoder produces the multi-scale
feature hierarchy
\begin{equation}
    \mathcal{E}_{\mathbf{x}}
    =
    \operatorname{Enc}(\mathbf{x})
    =
    \left\{
        \mathbf{E}_0,
        \mathbf{E}_1,
        \mathbf{E}_2,
        \mathbf{E}_3,
        \mathbf{E}_4,
        \mathbf{B}
    \right\},
\end{equation}
where $\mathbf{E}_0$--$\mathbf{E}_4$ denote the multi-scale encoder features
and $\mathbf{B}$ denotes the bottleneck representation. Together with the
partially decoded feature $\mathbf{D}_2$, they form the input to the
multi-scale observation module.
Each source feature is projected to a common channel dimensionality using a
$1\times1\times1$ convolution and resized to the spatial resolution of
$\mathbf{D}_2$. The aligned features are concatenated along the channel
dimension and fused as
\begin{equation}
    \mathbf{o}
    =
    \Phi_{\mathrm{fuse}}
    \left(
        \operatorname{Concat}
        \left\{
            \mathcal{R}_2
            \left(
                \phi_j(\mathbf{F}_j)
            \right)
            \,\middle|\,
            \mathbf{F}_j
            \in
            \mathcal{E}_{\mathbf{x}}
            \cup
            \left\{\mathbf{D}_2\right\}
        \right\}
    \right),
\end{equation}
where $\phi_j$ denotes the channel projection for the $j$-th source feature,
$\mathcal{R}_2$ denotes resizing to the stage-2 resolution, and
$\Phi_{\mathrm{fuse}}$ comprises a $3\times3\times3$ convolution followed by
a $1\times1\times1$ projection. The resulting observation $\mathbf{o}$
integrates shallow boundary cues, deep semantic information, and
decoder-conditioned context for subsequent latent anatomical refinement.

\subsection{Action-conditioned RSSM Latent Rollout}
\label{subsec:rssm_rollout}

The latent world model follows a recurrent state-space formulation
\cite{hafner2019planet}, in which the state at rollout step $t$ consists of a
deterministic recurrent state $\mathbf{h}_t$ and a stochastic latent state
$\mathbf{z}_t$. The multi-scale observation initializes both states as:
\begin{equation}
    \mathbf{h}_0=\phi_h(\mathbf{o}),
    \qquad
    q_{\phi}(\mathbf{z}_0\mid\mathbf{h}_0,\mathbf{o})
    =
    \mathcal{N}
    \left(
        \boldsymbol{\mu}^{q}_0,
        \operatorname{diag}\big[(\boldsymbol{\sigma}^{q}_0)^2\big]
    \right).
\end{equation}

During training, $\mathbf{z}_0$ is sampled from the posterior using the
reparameterization trick, whereas inference uses the posterior mean
$\boldsymbol{\mu}^{q}_0$.

The model then performs $T=3$ action-conditioned transitions. At each step,
the actor predicts hierarchical anatomical actions, which are provided to a
3D ConvGRU together with the previous stochastic state:
\begin{equation}
    \mathbf{u}_t
    =
    \mathcal{A}_{\theta}
    \left(
        [\mathbf{h}_{t-1},\mathbf{z}_{t-1}]
    \right),
    \qquad
    \widehat{\mathbf{a}}_t=\sigma(\mathbf{u}_t),
    \qquad
    \mathbf{h}_t
    =
    \operatorname{ConvGRU}
    \left(
        \mathbf{h}_{t-1},
        [\mathbf{z}_{t-1},\widehat{\mathbf{a}}_t]
    \right),
\end{equation}

where $[\cdot,\cdot]$ denotes channel-wise concatenation. The prior and
observation-conditioned posterior are parameterized as:
\begin{equation}
\begin{aligned}
    p_{\theta}(\mathbf{z}_t\mid\mathbf{h}_t)
    &=
    \mathcal{N}
    \left(
        \boldsymbol{\mu}^{p}_t,
        \operatorname{diag}\big[(\boldsymbol{\sigma}^{p}_t)^2\big]
    \right),
    \qquad
    q_{\phi}(\mathbf{z}_t\mid\mathbf{h}_t,\mathbf{o})
    =
    \mathcal{N}
    \left(
        \boldsymbol{\mu}^{q}_t,
        \operatorname{diag}\big[(\boldsymbol{\sigma}^{q}_t)^2\big]
    \right).
\end{aligned}
\end{equation}

During training, $\mathbf{z}_t$ is sampled from
$q_{\phi}(\mathbf{z}_t\mid\mathbf{h}_t,\mathbf{o})$. During inference, the
posterior is used only for initialization, and subsequent transitions use
\begin{equation}
    \mathbf{z}_t=\boldsymbol{\mu}^{p}_t,
    \qquad t=1,\ldots,T.
\end{equation}

The posterior is conditioned exclusively on image-derived features; the
ground-truth segmentation is used only to construct the training objectives
and is never provided as a posterior input.

A shared low-resolution segmentation head maps each latent state to atomic
logits. After the final transition, the deterministic and stochastic states
are projected back to the stage-2 decoder feature space:
\begin{equation}
    \boldsymbol{\ell}^{\mathrm{aux}}_t
    =
    \mathcal{H}_{\mathrm{aux}}
    \left(
        [\mathbf{h}_t,\mathbf{z}_t]
    \right),
    \qquad
    \widetilde{\mathbf{D}}_2
    =
    \phi_{\mathrm{final}}
    \left(
        [\mathbf{h}_T,\mathbf{z}_T]
    \right).
\end{equation}

The prediction at $t=0$ is used only to construct the target for the first
action, whereas the predictions after the recurrent transitions
$t=1,\ldots,T$ receive auxiliary segmentation supervision.
The refined representation $\widetilde{\mathbf{D}}_2$ remains a latent feature
and is subsequently processed by the preserved high-resolution decoder to
produce the final segmentation.

\subsection{Foreground-masked Balanced Hierarchical Action Learning}
\label{subsec:balanced_action}

\subsubsection{Hierarchical Action Targets}

At each recurrent transition, the actor learns anatomical corrections between
the current low-resolution prediction and the ground truth. We define
$G=73$ anatomical groups, comprising 35 atomic groups, 35 canonical groups,
and three global groups representing the complete foreground, cartilage, and
non-cartilage regions. A fixed atomic-to-group mapping is used to aggregate
both the predicted atomic probabilities and the ground-truth labels.

Let $Y_{b,g}(v)\in\{0,1\}$ denote the ground-truth mask of group $g$ for
sample $b$. Let $P_{b,t,g}(v)\in[0,1]$ denote the corresponding group
probability derived from the auxiliary prediction at the source state
$[\mathbf{h}_{t-1},\mathbf{z}_{t-1}]$ of transition $t$. The probabilities
used for target construction are detached from the computational graph.
The soft add and remove targets are defined as
\begin{equation}
    A^{\mathrm{add}}_{b,t,g}(v)
    =
    Y_{b,g}(v)\bigl[1-P_{b,t,g}(v)\bigr],
    \qquad
    A^{\mathrm{remove}}_{b,t,g}(v)
    =
    \bigl[1-Y_{b,g}(v)\bigr]P_{b,t,g}(v).
\end{equation}

The add target emphasizes ground-truth regions missing from the current
prediction, whereas the remove target emphasizes predictions inconsistent
with the ground truth. Because the targets are derived from probabilities,
they represent continuous correction magnitudes rather than binary error
masks.

\subsubsection{Balanced Foreground-masked Action Objective}

Let $s\in\{\mathrm{add},\mathrm{remove}\}$ denote the correction type, with
\begin{equation}
    A_{b,t,g,\mathrm{add}}
    =
    A^{\mathrm{add}}_{b,t,g},
    \qquad
    A_{b,t,g,\mathrm{remove}}
    =
    A^{\mathrm{remove}}_{b,t,g}.
\end{equation}

The actor output $\widehat{\mathbf{a}}_t$ contains the predicted action maps
$\widehat{A}_{b,t,g,s}(v)\in[0,1]$ for all anatomical groups and correction
types.

To balance the anatomical hierarchy, total prior masses of $0.6$, $0.3$, and
$0.1$ are allocated to the atomic, canonical, and global levels,
respectively, and distributed uniformly among the groups within each level.
The resulting group prior is denoted by $\pi_g$. For sample $b$, the presence
coefficient $q_{b,g}$ is set to $1$ when group $g$ is present in the ground
truth and to $0.1$ otherwise. Add and remove corrections receive coefficients
$\lambda_{\mathrm{add}}=2$ and $\lambda_{\mathrm{remove}}=1$, respectively.
The normalized channel weight is
\begin{equation}
    w_{b,g,s}
    =
    \frac{
        \pi_g q_{b,g}\lambda_s
    }{
        \displaystyle
        \sum_{g'=1}^{G}
        \sum_{s'\in\{\mathrm{add},\mathrm{remove}\}}
        \pi_{g'}q_{b,g'}\lambda_{s'}
        +\epsilon
    },
    \qquad
    s\in\{\mathrm{add},\mathrm{remove}\},
\end{equation}

where $\epsilon$ is a small constant for numerical stability. These channel
weights are computed for each training patch and shared across its recurrent
transitions.

Action supervision is further concentrated on foreground-relevant regions.
The ground-truth foreground is dilated once using a
$3\times3\times3$ kernel and combined with the confidently predicted
foreground. A voxel belongs to the predicted foreground region
$\widehat{\Omega}^{\mathrm{fg}}_{b,t}$ when its maximum foreground probability
exceeds both the background probability and $0.5$. We assign the spatial
weight $\rho_{b,t}(v)=1$ inside the union of the dilated ground-truth
foreground and $\widehat{\Omega}^{\mathrm{fg}}_{b,t}$, and
$\rho_{b,t}(v)=0.02$ elsewhere. Thus, background voxels are down-weighted rather than completely excluded,
preserving supervision for false-positive removal. At the patch level, let
$r_b=|\Omega|^{-1}\sum_{v\in\Omega}Y_b^{\mathrm{fg}}(v)$ denote the
foreground fraction, where $\Omega$ is the patch domain. The patch coefficient
$\psi(r_b)$ is set to $1$ when $r_b>10^{-5}$ and to $0.1$ otherwise, thereby
reducing the contribution of background-only patches.

Let $\ell_{\mathrm{act}}(\widehat{A},A;\rho)$ denote the sum of spatially
weighted binary cross-entropy and soft Dice losses. Within the binary
cross-entropy term, the positive contribution at voxel $v$ is additionally
weighted by $1+4A(v)$, assigning greater importance to locations requiring
stronger corrections. The complete action objective is
\begin{equation}
    \mathcal{L}_{\mathrm{action}}
    =
    \frac{1}{TB}
    \sum_{t=1}^{T}
    \sum_{b=1}^{B}
    \psi(r_b)
    \sum_{g=1}^{G}
    \sum_{s\in\{\mathrm{add},\mathrm{remove}\}}
    w_{b,g,s}\,
    \ell_{\mathrm{act}}
    \left(
        \widehat{A}_{b,t,g,s},
        A_{b,t,g,s};
        \rho_{b,t}
    \right).
\end{equation}

The channel weights balance anatomical hierarchy levels, group presence, and
correction types, whereas the spatial and patch-level weights reduce
domination by extensive background regions and background-only patches.

\subsection{Training Objective and Inference}
\label{subsec:training_inference}

The final full-resolution prediction and the recurrent auxiliary predictions
are supervised using equally weighted cross-entropy and soft Dice losses.
Let $\mathcal{L}_{\mathrm{final}}$ denote the final segmentation loss. The
auxiliary segmentation and KL-divergence objectives are averaged over the
$T$ recurrent transitions:
\begin{equation}
    \mathcal{L}_{\mathrm{aux}}
    =
    \frac{1}{T}
    \sum_{t=1}^{T}
    \mathcal{L}^{t}_{\mathrm{seg}},
    \qquad
    \mathcal{L}_{\mathrm{KL}}
    =
    \frac{1}{T}
    \sum_{t=1}^{T}
    D_{\mathrm{KL}}
    \left(
        q_{\phi}(\mathbf{z}_t\mid\mathbf{h}_t,\mathbf{o})
        \,\|\, 
        p_{\theta}(\mathbf{z}_t\mid\mathbf{h}_t)
    \right).
\end{equation}

The complete training objective is
\begin{equation}
    \mathcal{L}
    =
    \mathcal{L}_{\mathrm{final}}
    +
    0.2\,\mathcal{L}_{\mathrm{aux}}
    +
    0.1\,\mathcal{L}_{\mathrm{KL}}
    +
    0.2\,\mathcal{L}_{\mathrm{action}}.
\end{equation}

During training, stochastic states are sampled from the
observation-conditioned posterior at each transition. At inference, the
posterior mean initializes $\mathbf{z}_0$, and the subsequent $T=3$
transitions use the corresponding prior means. The final latent state is
projected into the preserved high-resolution decoder and combined with
$\mathbf{E}_1$ and $\mathbf{E}_0$ to produce the full-resolution atomic
segmentation. No ground-truth labels or action targets are required during
inference.

\section{Experiments}
\label{sec:experiments}

\subsection{Dataset and Annotation}
\label{sec:dataset}

We evaluated the proposed method on an in-house head CT dataset for
fine-grained auricular anatomy segmentation. The dataset comprised 202
three-dimensional CT volumes, of which 121 cases were used for training,
41 for validation, and 32 for testing. The same split
was used for all methods, and model selection was performed exclusively on
the validation set.

The original annotations describe nested anatomical structures: a voxel may
belong simultaneously to a local auricular substructure, its cartilage
counterpart, and a larger parent structure. Since conventional multi-class
segmentation requires mutually exclusive targets, we converted the overlapping
annotations into 35 non-overlapping foreground labels, referred to as
\emph{atomic labels}, together with the background class. Parent residual
regions were explicitly retained during this conversion. Consequently, the
atomic labels form a complete partition of the annotated foreground while
preserving the information required to recover the original hierarchy.

For evaluation, the atomic predictions were deterministically recomposed into
35 canonical anatomical structures according to a predefined inclusion
dictionary. Specifically, the mask of canonical structure $c$ was obtained as

\begin{equation}
    Y^{\mathrm{can}}_c
    =
    \bigcup_{k\in\mathcal{A}(c)}
    Y^{\mathrm{atom}}_k,
\end{equation}

where $\mathcal{A}(c)$ denotes the set of atomic labels belonging to canonical
structure $c$. All compared networks were therefore optimized using the same
mutually exclusive atomic targets, whereas the reported anatomical performance
was measured after recovering the original canonical structures.

\subsection{Pre-processing and Data Augmentation}
\label{sec:preprocessing}

AuricularWorld and all baselines except UNETR and SwinUNETR used the
automatically configured nnU-Net preprocessing and augmentation
pipeline~\cite{isensee2021nnunet}. Volumes were resampled to
$0.500 \times 0.429 \times 0.429\ \mathrm{mm}^{3}$ and trained using
$128^3$ patches. Intensities were clipped to $[-644,243]$ HU and normalized
using a training-set foreground mean and standard deviation of $-28.808$ and
$138.011$ HU, respectively. The default nnU-Net augmentations and a foreground
oversampling probability of $0.33$ were used, without additional
region-of-interest cropping.

For UNETR and SwinUNETR, we adapted the official MONAI
pipelines~\cite{hatamizadeh2022unetr,hatamizadeh2022swin,cardoso2022monai}.
Raw volumes were reoriented to RAS, resampled to
$1.5 \times 1.5 \times 2.0\ \mathrm{mm}^{3}$, clipped to $[-175,250]$ HU,
and normalized to $[0,1]$. Training used zero-padded $96^3$ patches,
foreground--background balanced sampling, and the default spatial and
intensity augmentations. One patch was generated per sampled volume, with
250 samples drawn with replacement per epoch. UNETR determined the foreground
crop before normalization using a threshold of $-1000$ HU, whereas SwinUNETR
retained the MONAI nonzero-voxel crop after normalization. All methods used
the same patient split, atomic labels, and evaluation protocol.





\subsection{Implementation Details}
\label{sec:implementation_details}

For AuricularWorld, multi-scale features were aligned at a spatial resolution
of $32^3$ and fused into a 128-channel observation. The deterministic and
stochastic RSSM states contained 128 and 32 channels, respectively, and the
latent rollout comprised three recurrent transitions. The predicted log
variances were clamped to $[-20,2]$. Standard nnU-Net deep supervision was
replaced by auxiliary segmentation heads attached to the recurrent states.
These architectural settings were fixed across all ablation variants.

The loss formulation followed Section~\ref{sec:methodology}. The final
segmentation, recurrent auxiliary segmentation, KL-divergence, and
hierarchical action losses were weighted by $1.0$, $0.2$, $0.1$, and $0.2$,
respectively. Segmentation losses combined soft Dice and cross-entropy with
equal weights, auxiliary targets were downsampled using nearest-neighbor
interpolation, and the KL loss was averaged over the three recurrent
transitions with zero free bits.

Methods implemented using the nnU-Net training pipeline were trained for
500 epochs, with 250 training and 50 validation iterations per epoch and a
batch size of two. We used SGD with Nesterov momentum of $0.99$, an initial
learning rate of $10^{-2}$, weight decay of $3\times10^{-5}$, and the nnU-Net
polynomial learning-rate schedule. Automatic mixed precision was enabled, and
gradients were clipped to a maximum norm of 12. For UNETR and SwinUNETR, the
adapted MONAI pipelines generated one patch per sampled volume, with 250 volume
samples drawn with replacement in each epoch.

For each run, model selection was based exclusively on validation performance.
Each method was independently trained using the same three random seeds, with
the Python, NumPy, and PyTorch random-number generators initialized
accordingly. Results are reported as mean $\pm$ standard deviation across the
three runs. Experiments were conducted using Python 3.10, PyTorch 2.12.0, and
CUDA 12.6 on a single NVIDIA GeForce RTX 4090 GPU with 24 GB of memory.

\subsection{Comparison Methods and Ablation Settings}
\label{sec:comparison_methods}

We compared AuricularWorld with nnU-Net, TransUNet
\cite{chen2021transunet}, nnFormer \cite{zhou2023nnformer}, nnMamba
\cite{gong2025nnmamba}, UNETR \cite{hatamizadeh2022unetr}, and SwinUNETR
\cite{hatamizadeh2022swin}. All methods used the same patient-level split,
atomic label representation, and evaluation protocol. AuricularWorld,
nnU-Net, TransUNet, nnFormer, and nnMamba used the nnU-Net preprocessing and
training pipeline, whereas UNETR and SwinUNETR followed the adapted MONAI
pipelines described in Section~\ref{sec:preprocessing}. 

The ablation study progressively extended nnU-Net with the multi-scale action
RSSM (MS-RSSM), balanced action weighting (Bal.), and foreground-masked action
supervision (FG). MS-RSSM introduced multi-scale observation fusion, a
three-step recurrent latent rollout, and hierarchical add/remove action
supervision. Bal.\ added anatomical-group, group-presence, add/remove-channel,
and patch-level weighting, while FG spatially focused the action objective on
foreground-relevant regions. Their complete combination constitutes
AuricularWorld. All other architectural and loss settings were fixed across
the ablation variants.

\subsection{Evaluation Metrics}
\label{sec:evaluation_protocol}

Atomic predictions were obtained by voxel-wise argmax and deterministically recomposed into 35
canonical anatomical structures using the fixed atomic-to-canonical mapping.
No method-specific post-processing was applied.

Performance was evaluated using the Dice similarity coefficient and the
95th-percentile symmetric Hausdorff distance (HD95, in millimeters). HD95 was
computed from bidirectional surface distances in physical space using the
original image spacing. When both masks were empty, HD95 was set to zero; when
only one mask was empty, it was set to infinity. For each random seed,
structure-wise metrics were first averaged across the 32 eligible test cases
and then macro-averaged over the 35 canonical structures. Results are reported
as mean $\pm$ standard deviation across three random seeds. For the external
comparison, we additionally report the number of structures on which each
method achieved the best mean Dice or HD95.

\begin{table*}[t]
    \centering
    \caption{
    External comparison on the test set.
    Results are reported as mean $\pm$ standard deviation across three random
    seeds. ``Best'' denotes the number of canonical structures for which a
    method achieved the best mean score among all compared methods.
    }
    \label{tab:external_comparison}

    \small
    \setlength{\tabcolsep}{8pt}
    \begin{tabular}{@{}lcccc@{}}
        \toprule
        Method
        & Dice (\%) $\uparrow$
        & HD95 (mm) $\downarrow$
        & \makecell{Dice\\best}
        & \makecell{HD95\\best} \\
        \midrule

        nnU-Net~\cite{isensee2021nnunet}
        & $77.75 \pm 0.18$
        & $2.426 \pm 1.014$
        & 2
        & 3 \\

        TransUNet~\cite{chen2021transunet}
        & $77.75 \pm 0.31$
        & $2.466 \pm 1.683$
        & 2
        & 5 \\

        nnFormer~\cite{zhou2023nnformer}
        & $75.76 \pm 0.91$
        & $11.318 \pm 5.373$
        & 0
        & 0 \\

        SwinUNETR~\cite{hatamizadeh2022swin}
        & $66.45 \pm 0.38$
        & $\infty$
        & 0
        & 0 \\

        nnMamba~\cite{gong2025nnmamba}
        & $66.01 \pm 10.85$
        & $\infty$
        & 0
        & 1 \\

        UNETR~\cite{hatamizadeh2022unetr}
        & $56.86 \pm 2.08$
        & $\infty$
        & 0
        & 0 \\
        \midrule

        \textbf{AuricularWorld}
        & $\mathbf{78.42 \pm 0.09}$
        & $\mathbf{1.379 \pm 0.014}$
        & \textbf{31}
        & \textbf{26} \\

        \bottomrule
    \end{tabular}
\end{table*}

\subsection{Experimental Results}
\label{sec:experimental_results}

\subsubsection{Comparison with Existing Segmentation Methods}

Table~\ref{tab:external_comparison} compares AuricularWorld with representative
CNN-, Transformer-, and state-space-model-based segmentation methods on the
32 eligible test cases. AuricularWorld achieved the highest macro-average Dice
of $78.42\%$ and the lowest HD95 of $1.379$ mm. It also obtained the best mean
Dice on 31 of the 35 canonical structures and the best mean HD95 on 26
structures, demonstrating consistently strong performance across the
fine-grained auricular anatomy.

Compared with nnU-Net, AuricularWorld improved Dice by $0.67$ percentage points
and reduced HD95 from $2.426$ to $1.379$ mm, corresponding to a relative
reduction of approximately $43.2\%$. It similarly outperformed TransUNet by
$0.67$ percentage points in Dice and reduced its HD95 from $2.466$ to
$1.379$ mm, corresponding to a relative reduction of approximately $44.1\%$.
Among the methods with finite distance estimates, AuricularWorld also exhibited
the lowest variation across random seeds, particularly for HD95.

Although nnFormer maintained a moderate Dice score, it produced substantially
larger boundary errors. Under the predefined empty-mask convention, SwinUNETR,
nnMamba, and UNETR produced one-sided empty-mask predictions, resulting in
infinite aggregate HD95 values. Overall, AuricularWorld provided the best
balance of volumetric overlap, boundary accuracy, and training stability among
the compared methods.

\begin{table}[t]
    \centering
    \caption{
    Incremental ablation of AuricularWorld.
    MS-RSSM denotes the multi-scale action RSSM, Bal.\ denotes balanced
    action weighting, and FG denotes foreground-masked action supervision.
    }
    \label{tab:internal_comparison}

    \small
    \setlength{\tabcolsep}{5pt}
    \begin{tabular}{@{}lccccc@{}}
        \toprule
        Model
        & MS-RSSM
        & Bal.
        & FG
        & Dice (\%) $\uparrow$
        & HD95 (mm) $\downarrow$ \\
        \midrule

        nnU-Net
        & $\times$
        & $\times$
        & $\times$
        & $77.75 \pm 0.18$
        & $2.426 \pm 1.014$ \\

        $+$ MS-RSSM
        & $\checkmark$
        & $\times$
        & $\times$
        & $78.20 \pm 0.08$
        & $1.406 \pm 0.058$ \\

        $+$ Bal.
        & $\checkmark$
        & $\checkmark$
        & $\times$
        & $78.32 \pm 0.17$
        & $1.537 \pm 0.289$ \\

        \textbf{AuricularWorld}
        & $\checkmark$
        & $\checkmark$
        & $\checkmark$
        & $\mathbf{78.42 \pm 0.09}$
        & $\mathbf{1.379 \pm 0.014}$ \\

        \bottomrule
    \end{tabular}
\end{table}

\subsubsection{Incremental Component Ablation}

Table~\ref{tab:internal_comparison} presents an incremental ablation study
starting from the nnU-Net baseline. Adding the multi-scale action RSSM
(MS-RSSM) increased the mean Dice from $77.75\%$ to $78.20\%$ and reduced
HD95 from $2.426$ to $1.406$ mm, corresponding to a $0.45$ percentage-point
improvement in Dice and an approximately $42.0\%$ relative reduction in HD95.
This substantial decrease in boundary error demonstrates the effectiveness of
multi-scale recurrent latent refinement over the feed-forward baseline.

Introducing balanced action weighting further increased Dice to $78.32\%$,
but increased HD95 from $1.406$ to $1.537$ mm. This suggests that balancing
the hierarchical anatomical groups and add/remove action channels improved
overall volumetric overlap, but did not by itself provide more accurate
boundary localization.

The complete AuricularWorld model, obtained by additionally applying
foreground-masked action supervision, achieved the highest Dice of $78.42\%$
and the lowest HD95 of $1.379$ mm. Compared with the balanced-action
configuration, foreground masking improved Dice by a further $0.10$ percentage
points and reduced HD95 by approximately $10.3\%$. This result suggests that
spatially concentrating action supervision on foreground-relevant regions
helps suppress background-dominated corrections and improves both overlap and
boundary accuracy.

Overall, AuricularWorld improved Dice by $0.67$ percentage points and reduced
HD95 by approximately $43.2\%$ relative to nnU-Net. The incremental results
show that MS-RSSM provides the principal improvement over the baseline, while
balanced action weighting and foreground-masked supervision provide
complementary refinements that produce the best final performance.

\subsubsection{Qualitative Comparison}

\begin{figure*}[t]
    \centering
    \includegraphics[width=\textwidth]{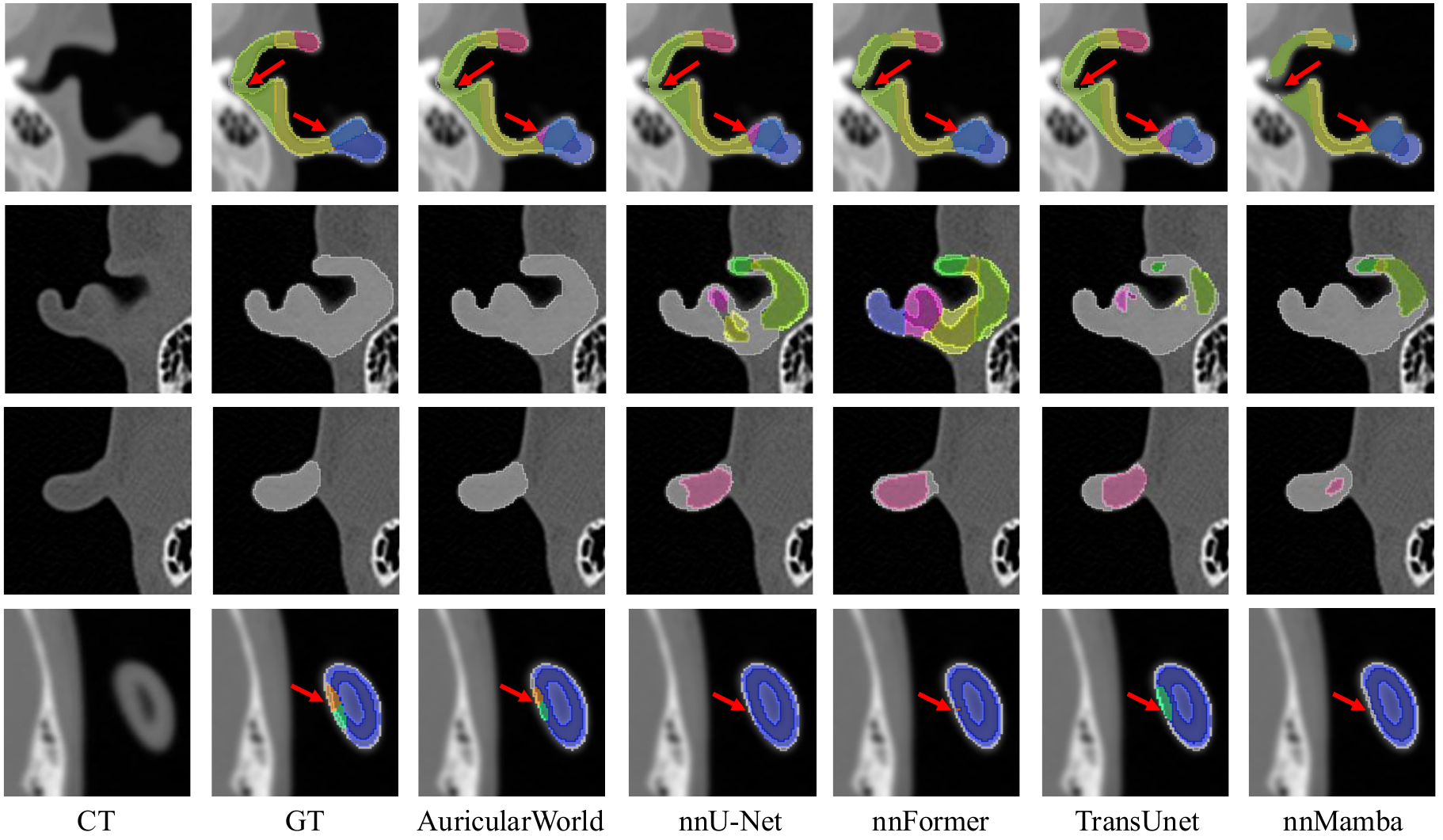}
    \caption{
    Qualitative comparison on representative test cases. Columns show the input
    CT, ground truth, and predictions from AuricularWorld, nnU-Net, nnFormer,
    TransUNet, and nnMamba. Colors denote canonical auricular structures, and red
    arrows highlight challenging regions. AuricularWorld better preserves
    structural continuity and boundary details while reducing missed and
    fragmented predictions.
    }
    \label{fig:qualitative_comparison}
\end{figure*}

Figure~\ref{fig:qualitative_comparison} presents representative qualitative
comparisons across challenging auricular structures. AuricularWorld produced
predictions that were generally more consistent with the ground truth in terms
of structural continuity, label assignment, and boundary delineation. In the
first example, AuricularWorld better preserved the narrow transitions between
adjacent anatomical regions while reducing boundary leakage. In the second and
third examples, it maintained more complete target shapes and avoided the
fragmentation and spurious internal labels observed in several competing
methods. In the final example, AuricularWorld recovered the thin peripheral
component highlighted by the red arrow, whereas the baseline methods either
missed, truncated, or over-expanded this region. These visual results support
the quantitative findings and suggest that multi-scale recurrent latent
refinement improves the delineation of small, thin, and closely adjacent
auricular structures.

\begin{figure*}[t]
    \centering
    \includegraphics[width=\textwidth]{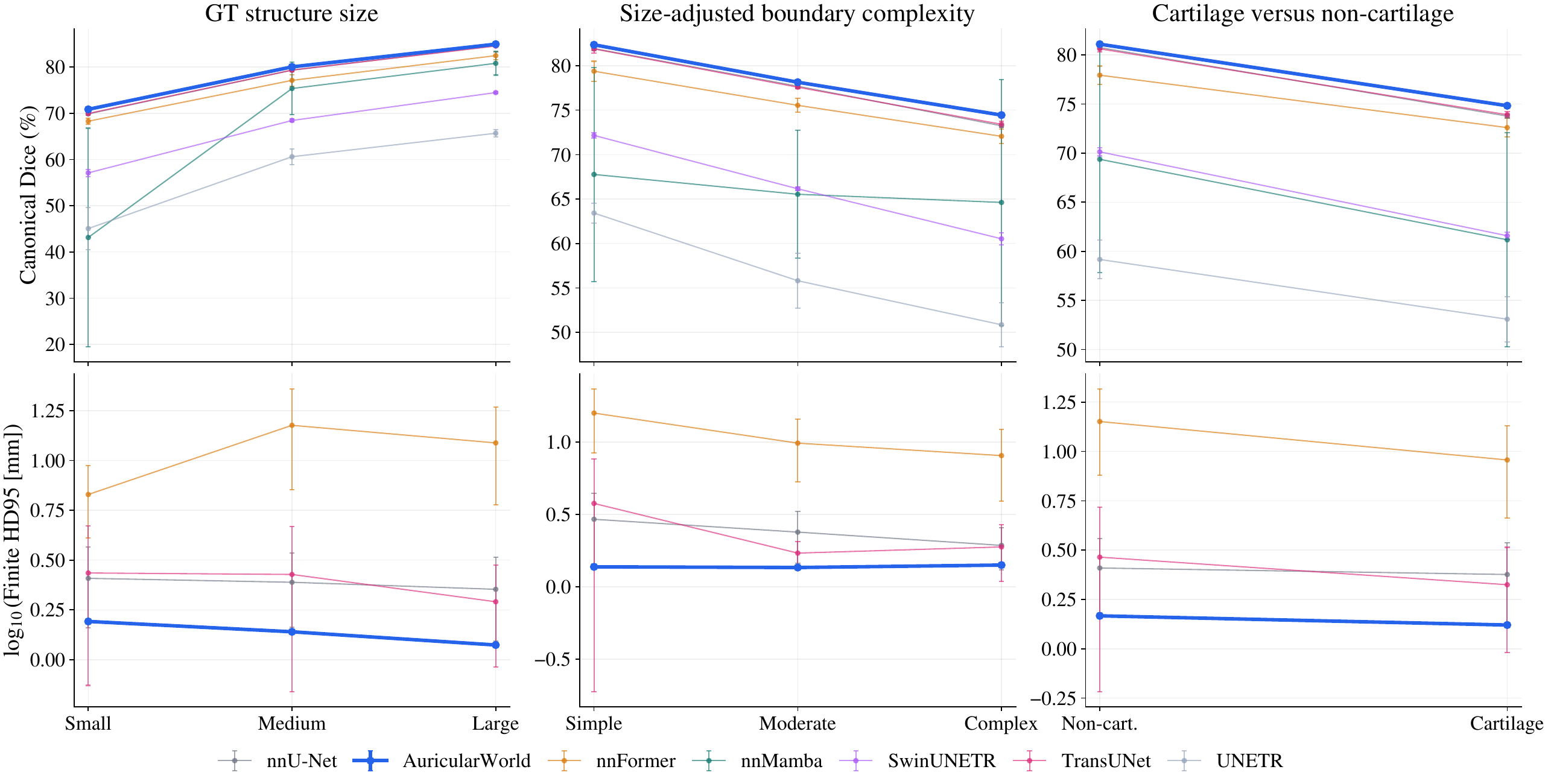}
    \caption{
    Category-wise comparison of AuricularWorld and external segmentation
    methods on test set. The columns stratify the anatomical
    structures according to ground-truth physical size, size-adjusted boundary
    complexity, and tissue type, respectively. The top row reports canonical
    Dice scores, while the bottom row reports finite $\mathrm{HD}_{95}$ values.
    Structure size was divided into tertiles based on median physical volume.
    Boundary complexity was defined using the residual surface area after
    removing its log-linear dependence on physical volume. The cartilage and
    non-cartilage groups each contained 17 structures; the whole auricle was
    excluded from the tissue comparison because it contains both tissue types.
    Points indicate the mean over three random seeds, and error bars denote one standard
    deviation. Higher Dice and lower $\mathrm{HD}_{95}$ indicate better
    performance.
    }
    \label{fig:category_analysis}
\end{figure*}

\subsubsection{Performance across anatomical structure categories.}
We further evaluated segmentation performance according to structure size,
boundary complexity, and tissue type, as shown in
Fig.~\ref{fig:category_analysis}. Structure size was determined using tertiles
of the median physical volume measured from the ground-truth masks. To avoid
confounding boundary complexity with structure size, we regressed the
logarithm of surface area against the logarithm of volume and used the
resulting residuals to divide the structures into simple, moderate, and
complex boundary groups. Cartilage and non-cartilage groups each contained
17 structures, while the whole auricle was excluded from this comparison
because it contains both tissue types. 

AuricularWorld achieved the highest Dice score and the lowest finite
$\mathrm{HD}_{95}$ in all evaluated categories. Its advantage was particularly
pronounced for small structures, complex boundaries, and cartilage
structures. Specifically, AuricularWorld obtained Dice scores of $70.84\%$,
$74.45\%$, and $74.82\%$ for these three categories, respectively,
outperforming the strongest external method by $0.90$, $1.03$, and $0.91$
percentage points. The corresponding finite $\mathrm{HD95}$ values were
$1.56$, $1.41$, and $1.32$~mm. The larger improvement observed for complex
boundaries and cartilage structures suggests that foreground-masked action
learning primarily benefits thin, irregular, and spatially fragmented
anatomical regions rather than only improving large and geometrically simple

\begin{table*}[t]
\centering
\caption{Structure-wise comparison between nnU-Net and AuricularWorld for
non-cartilage and global anatomical structures.
Results are reported as mean $\pm$ standard deviation across three random
seeds.}
\label{tab:structure_wise_non_cartilage}
\scriptsize
\setlength{\tabcolsep}{4pt}
\renewcommand{\arraystretch}{1.03}
\resizebox{\textwidth}{!}{%
\begin{tabular}{lcccc}
\toprule
& \multicolumn{2}{c}{Dice (\%) $\uparrow$}
& \multicolumn{2}{c}{HD95 (mm) $\downarrow$} \\
\cmidrule(lr){2-3} \cmidrule(lr){4-5}
Structure & nnU-Net & AuricularWorld & nnU-Net & AuricularWorld \\
\midrule
Triangular Fossa & 70.69 $\pm$ 0.18 & \textbf{71.13 $\pm$ 0.29} & 1.917 $\pm$ 0.008 & \textbf{1.893 $\pm$ 0.030} \\
Antihelix Crus Residual & 88.65 $\pm$ 0.23 & \textbf{88.73 $\pm$ 0.05} & 1.255 $\pm$ 0.013 & \textbf{1.251 $\pm$ 0.020} \\
External Auditory Canal & \textbf{85.13 $\pm$ 0.75} & 85.11 $\pm$ 0.08 & 5.999 $\pm$ 2.075 & \textbf{1.181 $\pm$ 0.009} \\
Antitragus & 83.56 $\pm$ 0.29 & \textbf{83.62 $\pm$ 0.30} & 3.060 $\pm$ 2.694 & \textbf{1.558 $\pm$ 0.041} \\
Antihelix & 88.85 $\pm$ 0.19 & \textbf{88.92 $\pm$ 0.04} & \textbf{1.133 $\pm$ 0.006} & 1.146 $\pm$ 0.014 \\
Superior Crus Of Antihelix & 73.04 $\pm$ 0.16 & \textbf{73.57 $\pm$ 0.06} & 1.577 $\pm$ 0.057 & \textbf{1.525 $\pm$ 0.031} \\
Inferior Crus Of Antihelix & \textbf{72.64 $\pm$ 0.52} & 72.56 $\pm$ 0.50 & \textbf{1.336 $\pm$ 0.013} & 1.341 $\pm$ 0.013 \\
Intertragic Notch & 76.08 $\pm$ 0.71 & \textbf{77.15 $\pm$ 0.17} & 5.804 $\pm$ 4.301 & \textbf{1.388 $\pm$ 0.034} \\
Tragus & 82.77 $\pm$ 0.29 & \textbf{83.44 $\pm$ 0.54} & 2.884 $\pm$ 2.650 & \textbf{1.324 $\pm$ 0.052} \\
Concha & 80.30 $\pm$ 0.30 & \textbf{80.33 $\pm$ 0.04} & 2.962 $\pm$ 2.551 & \textbf{1.499 $\pm$ 0.020} \\
Cymba Conchae & 79.13 $\pm$ 0.33 & \textbf{79.22 $\pm$ 0.05} & 3.065 $\pm$ 2.557 & \textbf{1.577 $\pm$ 0.029} \\
Cavum Conchae & 80.72 $\pm$ 0.25 & \textbf{80.94 $\pm$ 0.23} & 1.617 $\pm$ 0.041 & \textbf{1.595 $\pm$ 0.051} \\
Scapha & 61.01 $\pm$ 0.69 & \textbf{62.76 $\pm$ 0.51} & 2.420 $\pm$ 0.081 & \textbf{2.390 $\pm$ 0.030} \\
Helix & 88.30 $\pm$ 0.06 & \textbf{88.41 $\pm$ 0.28} & 1.136 $\pm$ 0.011 & \textbf{1.127 $\pm$ 0.023} \\
Upper Helix & 88.61 $\pm$ 0.14 & \textbf{88.80 $\pm$ 0.42} & 1.127 $\pm$ 0.036 & \textbf{1.104 $\pm$ 0.046} \\
Crus Of Helix & 83.51 $\pm$ 0.30 & \textbf{83.63 $\pm$ 0.10} & 1.723 $\pm$ 0.054 & \textbf{1.712 $\pm$ 0.033} \\
Auricle & \textbf{94.65 $\pm$ 0.04} & 94.52 $\pm$ 0.15 & \textbf{0.810 $\pm$ 0.005} & 0.844 $\pm$ 0.021 \\
Ear Lobule & 89.50 $\pm$ 0.42 & \textbf{90.11 $\pm$ 0.24} & 4.618 $\pm$ 2.805 & \textbf{1.364 $\pm$ 0.052} \\
\bottomrule
\end{tabular}%
}
\end{table*}

\begin{table*}[t]
\centering
\caption{Structure-wise comparison between nnU-Net and AuricularWorld for
cartilage structures.
Results are reported as mean $\pm$ standard deviation across three random
seeds.}
\label{tab:structure_wise_cartilage}
\scriptsize
\setlength{\tabcolsep}{4pt}
\renewcommand{\arraystretch}{1.03}
\resizebox{\textwidth}{!}{%
\begin{tabular}{lcccc}
\toprule
& \multicolumn{2}{c}{Dice (\%) $\uparrow$}
& \multicolumn{2}{c}{HD95 (mm) $\downarrow$} \\
\cmidrule(lr){2-3} \cmidrule(lr){4-5}
Structure & nnU-Net & AuricularWorld & nnU-Net & AuricularWorld \\
\midrule
Cartilage Triangular Fossa & 67.38 $\pm$ 0.19 & \textbf{68.24 $\pm$ 0.34} & 1.241 $\pm$ 0.027 & \textbf{1.207 $\pm$ 0.020} \\
Cartilage Antihelix Crus Residual & 79.71 $\pm$ 0.23 & \textbf{80.63 $\pm$ 0.41} & 1.141 $\pm$ 0.004 & \textbf{1.128 $\pm$ 0.028} \\
Cartilage External Auditory Canal & 78.06 $\pm$ 0.40 & \textbf{78.51 $\pm$ 0.11} & 6.068 $\pm$ 2.076 & \textbf{1.209 $\pm$ 0.003} \\
Cartilage Antitragus & 76.48 $\pm$ 0.52 & \textbf{77.50 $\pm$ 0.57} & 2.888 $\pm$ 2.656 & \textbf{1.398 $\pm$ 0.042} \\
Cartilage Antihelix & 80.30 $\pm$ 0.23 & \textbf{81.15 $\pm$ 0.34} & 1.042 $\pm$ 0.014 & \textbf{1.034 $\pm$ 0.016} \\
Cartilage Superior Crus Of Antihelix & 66.74 $\pm$ 0.37 & \textbf{68.00 $\pm$ 0.12} & 1.494 $\pm$ 0.052 & \textbf{1.436 $\pm$ 0.019} \\
Cartilage Inferior Crus Of Antihelix & 69.50 $\pm$ 0.36 & \textbf{69.65 $\pm$ 0.35} & \textbf{1.294 $\pm$ 0.008} & 1.302 $\pm$ 0.016 \\
Cartilage Intertragic Notch & 67.11 $\pm$ 1.02 & \textbf{68.85 $\pm$ 0.50} & 5.683 $\pm$ 4.236 & \textbf{1.327 $\pm$ 0.036} \\
Cartilage Tragus & 75.27 $\pm$ 0.29 & \textbf{76.55 $\pm$ 0.17} & 2.753 $\pm$ 2.598 & \textbf{1.198 $\pm$ 0.044} \\
Cartilage Concha & 74.41 $\pm$ 0.37 & \textbf{75.09 $\pm$ 0.09} & 2.816 $\pm$ 2.547 & \textbf{1.348 $\pm$ 0.029} \\
Cartilage Cymba Conchae & 73.44 $\pm$ 0.34 & \textbf{74.02 $\pm$ 0.23} & 2.909 $\pm$ 2.550 & \textbf{1.431 $\pm$ 0.010} \\
Cartilage Cavum Conchae & 75.17 $\pm$ 0.53 & \textbf{76.30 $\pm$ 0.18} & 1.403 $\pm$ 0.043 & \textbf{1.371 $\pm$ 0.042} \\
Cartilage Scapha & 56.08 $\pm$ 0.45 & \textbf{58.23 $\pm$ 0.32} & \textbf{2.322 $\pm$ 0.071} & 2.335 $\pm$ 0.043 \\
Cartilage Helix & 77.33 $\pm$ 0.28 & \textbf{78.58 $\pm$ 0.11} & 1.204 $\pm$ 0.024 & \textbf{1.163 $\pm$ 0.019} \\
Cartilage Upper Helix & 75.83 $\pm$ 0.36 & \textbf{77.49 $\pm$ 0.25} & 1.168 $\pm$ 0.045 & \textbf{1.090 $\pm$ 0.030} \\
Cartilage Crus Of Helix & 75.56 $\pm$ 0.26 & \textbf{76.46 $\pm$ 0.16} & 1.723 $\pm$ 0.084 & \textbf{1.672 $\pm$ 0.017} \\
Auricular Cartilage & 85.64 $\pm$ 0.17 & \textbf{86.64 $\pm$ 0.14} & 3.305 $\pm$ 2.167 & \textbf{0.791 $\pm$ 0.008} \\
\bottomrule
\end{tabular}%
}
\end{table*}

\subsubsection{Structure-wise comparison.}
As shown in Table~\ref{tab:structure_wise_non_cartilage} and Table~\ref{tab:structure_wise_cartilage},
AuricularWorld achieves higher Dice scores than nnU-Net in 32 of the 35
anatomical structures and lower HD95 values in 30 structures. The largest
Dice improvements are observed for the cartilaginous scapha
($56.08\% \rightarrow 58.23\%$), scapha
($61.01\% \rightarrow 62.76\%$), cartilaginous intertragic notch
($67.11\% \rightarrow 68.85\%$), and cartilaginous upper helix
($75.83\% \rightarrow 77.49\%$). More substantial improvements are observed
in boundary accuracy. In particular, HD95 decreases from
$6.00$ to $1.18\,\mathrm{mm}$ for the external auditory canal,
from $6.07$ to $1.21\,\mathrm{mm}$ for its cartilaginous counterpart,
from $5.80$ to $1.39\,\mathrm{mm}$ for the intertragic notch, and from
$3.31$ to $0.79\,\mathrm{mm}$ for the overall auricular cartilage.
These results suggest that the proposed latent action-guided refinement is
particularly effective for thin cartilaginous structures and anatomically
complex boundaries. The few reductions in Dice are small, with the largest
being $0.14$ percentage points for the auricle, while the largest HD95
increase is only $0.034\,\mathrm{mm}$.

\begin{figure*}[t]
    \centering
    \includegraphics[width=\textwidth]{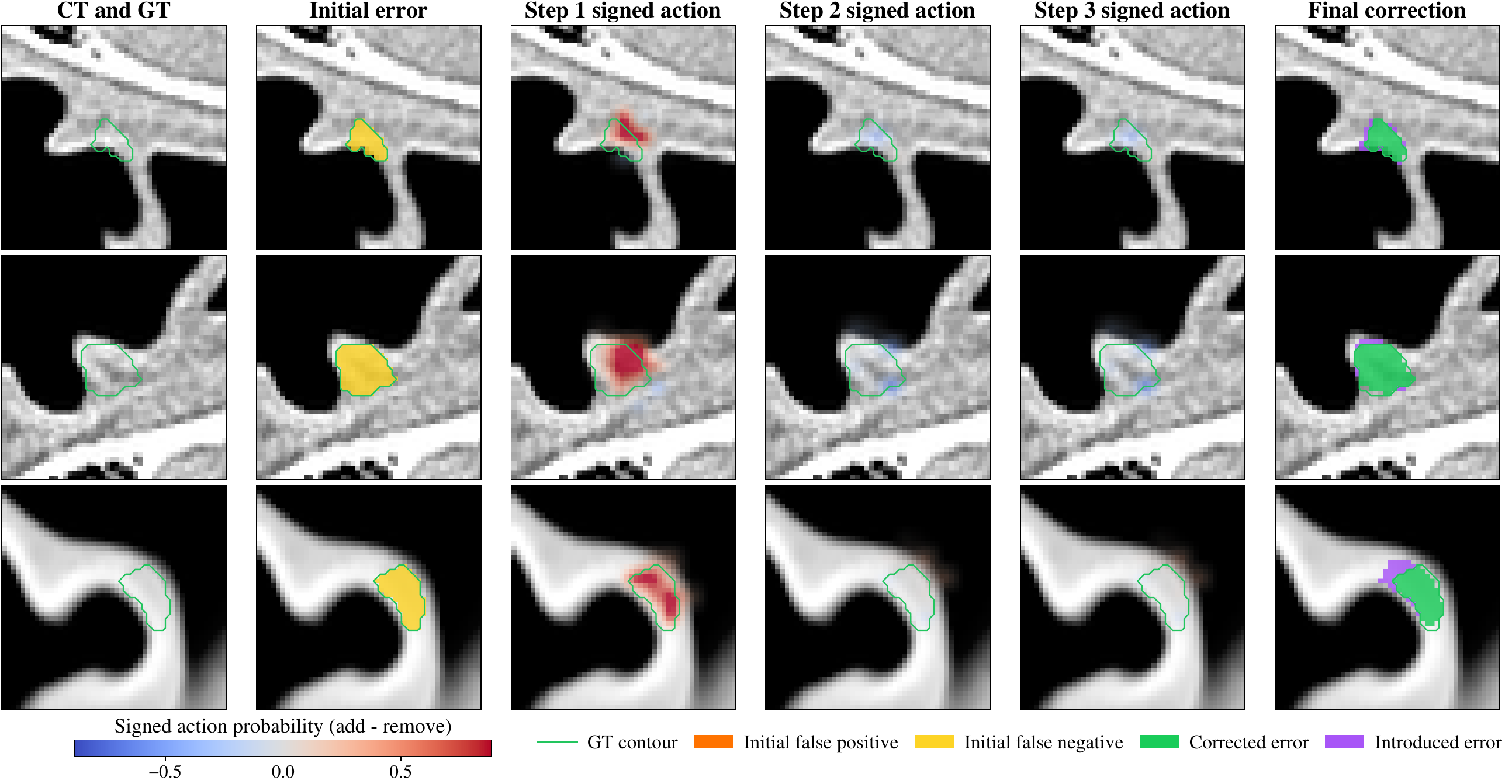}
    \caption{
    Visualization of the three-step RSSM action-guided refinement process.
    Each row shows one representative anatomical structure: the triangular
    fossa, inferior crus of the antihelix, and cartilaginous superior crus of
    the antihelix. From left to right, the columns show the CT image with the
    ground-truth contour, errors in the initial low-resolution state, signed
    actions at the three rollout steps, and the changes observed in the final
    high-resolution segmentation. The signed action is defined as the
    difference between the add and remove probabilities. Red and blue denote
    positive and negative actions, respectively. Orange and yellow indicate
    initial false-positive and false-negative regions, while green and purple
    indicate corrected and newly introduced errors, respectively. The green
    contour represents the ground truth. Positive actions at the first step
    are concentrated around initially missed structures, whereas subsequent
    actions primarily regulate their spatial extent and boundaries.
    }
    \label{fig:rssm_action_maps}
\end{figure*}

\subsubsection{Visualization of iterative action-guided refinement.}
Figure~\ref{fig:rssm_action_maps} visualizes the spatial actions generated during the three-step RSSM rollout. The three examples correspond to the triangular fossa, inferior crus of the antihelix, and cartilaginous superior crus of the antihelix. Before the first transition, all three structures are substantially under-segmented, as indicated by the yellow false-negative regions. At Step~1, strong positive actions are concentrated within or near the missing anatomical regions, increasing their latent responses. The actions become weaker and partially negative during Steps~2 and~3, suggesting that the model shifts from recovering the missing structure to suppressing excessive responses and stabilizing its boundary.

After the final RSSM state is passed through the high-resolution decoder, most initial errors are corrected, as shown by the green regions in the last column. The remaining purple regions indicate errors introduced during refinement and are primarily located near ambiguous boundaries. Overall, the action responses are spatially localized around the target structures and their initial errors rather than being distributed uniformly over the image. This behavior is consistent with the intended role of the action module: first recovering missing anatomical evidence and subsequently regularizing the refined latent representation.
structures.

\subsubsection{Hyperparameter sensitivity.}
\begin{figure}[t]
    \centering
    \includegraphics[width=\linewidth]
    {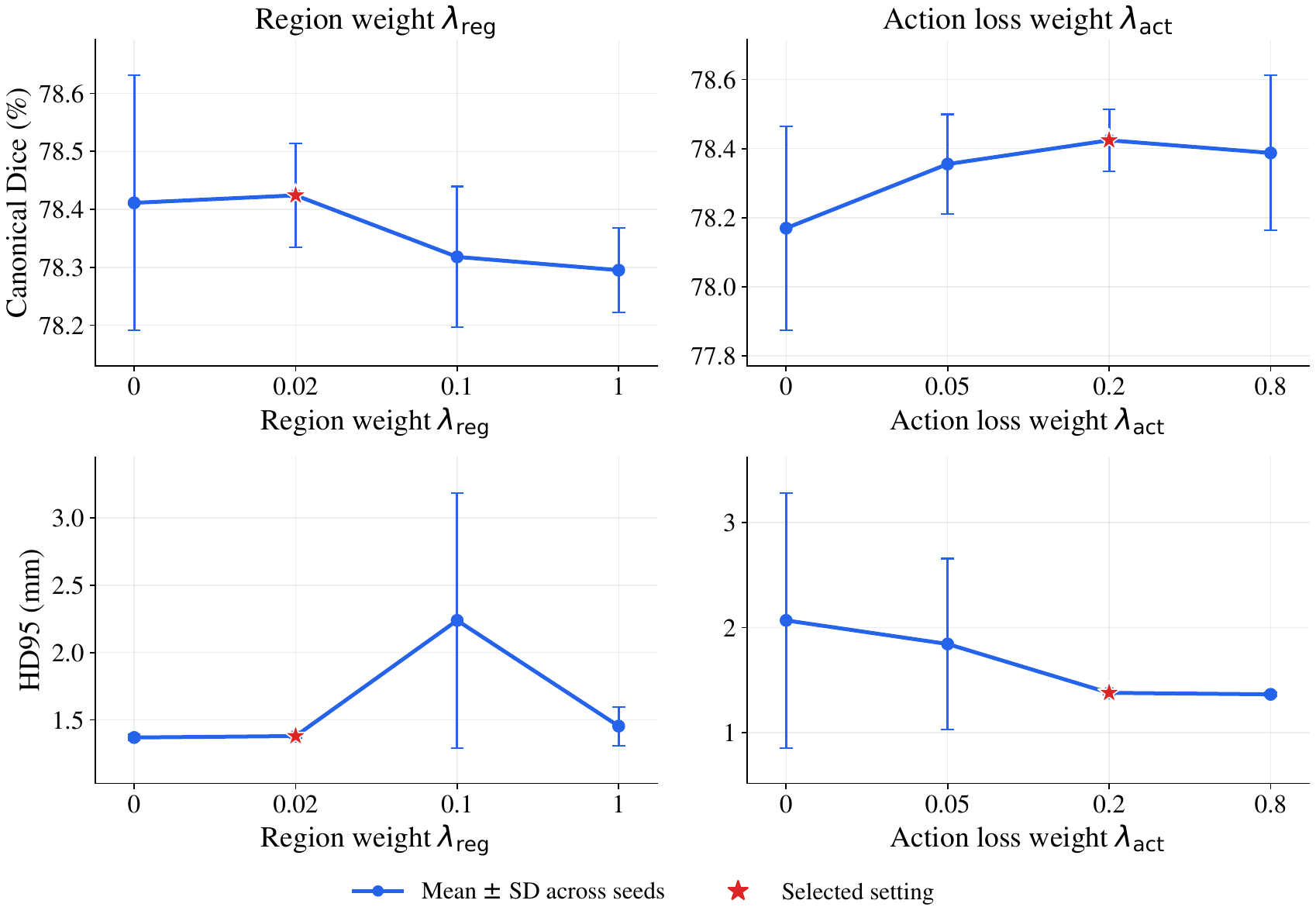}
    \caption{
    Sensitivity of AuricularWorld to the region weighting coefficient
    $\lambda_{\mathrm{reg}}$ and action supervision coefficient
    $\lambda_{\mathrm{act}}$. The upper and lower rows report canonical Dice
    and HD95, respectively. Each point represents the mean over three random
    seeds, and the error bars denote the corresponding standard deviation. Red stars indicate
    the selected default settings, namely
    $\lambda_{\mathrm{reg}}=0.02$ and
    $\lambda_{\mathrm{act}}=0.2$.
    }
    \label{fig:hyperparameter_sensitivity}
\end{figure}

We investigated the sensitivity of AuricularWorld to the region weighting
coefficient $\lambda_{\mathrm{reg}}$ and the action supervision coefficient
$\lambda_{\mathrm{act}}$. All results were averaged over the 35 canonical
structures. As shown in
Figure~\ref{fig:hyperparameter_sensitivity}, the segmentation performance was
relatively insensitive to small values of $\lambda_{\mathrm{reg}}$. The
default value $\lambda_{\mathrm{reg}}=0.02$ achieved the highest mean Dice of
$78.424\%$ with a small standard deviation of $0.090\%$. Increasing the region
weight to $0.1$ or $1.0$ slightly reduced Dice and resulted in less stable
HD95, particularly at $\lambda_{\mathrm{reg}}=0.1$.

The action supervision coefficient had a more pronounced effect. Removing
action supervision ($\lambda_{\mathrm{act}}=0$) reduced Dice to $78.170\%$
and increased HD95 to $2.068$~mm, with substantial variation across random
seeds. Increasing $\lambda_{\mathrm{act}}$ progressively improved both
metrics and reduced their variability. The default setting
$\lambda_{\mathrm{act}}=0.2$ produced the highest Dice of $78.424\%$ and a
stable HD95 of $1.379$~mm. Although $\lambda_{\mathrm{act}}=0.8$ achieved a
slightly lower mean HD95 of $1.365$~mm, it resulted in a lower Dice and
greater inter-seed variation. We therefore selected
$\lambda_{\mathrm{reg}}=0.02$ and $\lambda_{\mathrm{act}}=0.2$ as the final
configuration, providing the best overall trade-off between overlap accuracy,
boundary accuracy, and training stability.

\section{Conclusion}
\label{sec:conclusion}

We presented \emph{AuricularWorld}, a recurrent latent world-model framework
for fine-grained 3D auricular structure segmentation in CT. Built upon
nnU-Net, AuricularWorld integrates multi-scale encoder and decoder features
into an anatomical observation and performs a three-step RSSM rollout to
refine the intermediate latent representation. Hierarchical add/remove action
learning further guides the recurrent refinement, while balanced channel
weighting and foreground-masked supervision reduce the influence of sparse
structures and extensive background regions. The refined latent feature is
subsequently decoded using the preserved high-resolution skip connections,
allowing detailed anatomical boundaries to be reconstructed.

On the held-out test set, AuricularWorld achieved a Dice score of
$78.42 \pm 0.09\%$ and an HD95 of $1.379 \pm 0.014$ mm, outperforming the
evaluated CNN-, Transformer-, and state-space-model-based baselines. Compared
with nnU-Net, it improved Dice by $0.67$ percentage points and reduced HD95 by
approximately $43.2\%$. The incremental ablation study further showed that
multi-scale recurrent latent refinement provided the principal improvement,
while balanced action weighting and foreground-masked supervision contributed
complementary refinements. These results demonstrate the potential of latent
world-model reasoning for improving the segmentation of small, thin, and
closely adjacent auricular structures.

\bibliographystyle{elsarticle-num}
\bibliography{references}
\end{document}